\documentclass[sigconf,authorversion,nonacm]{acmart}
\usepackage{subcaption}
\usepackage{multicol}
\usepackage{soul} % Highlight using \hl{}
\AtBeginDocument{%
  \providecommand\BibTeX{{%
    \normalfont B\kern-0.5em{\scshape i\kern-0.25em b}\kern-0.8em\TeX}}}

\begin{document}
\fancyhead{}

\title{AlphaEvolve: A Learning Framework to Discover Novel Alphas in Quantitative Investment}

\author{Can Cui}
\affiliation{%
  \department{School of Computing}
  \institution{National University of Singapore}
}
\email{cuican@comp.nus.edu.sg}

\author{Wei Wang}
\affiliation{%
  \department{School of Computing}
  \institution{National University of Singapore}
}
\email{wangwei@comp.nus.edu.sg}

\author{Meihui Zhang}
\affiliation{%
  \department{School of Computer Science and Technology}
  \institution{Beijing Institute of Technology}
}
\email{meihui_zhang@bit.edu.cn}

\author{Gang Chen}
\affiliation{%
  \department{College of Computer Science and Technology}
  \institution{Zhejiang University}
}
\email{cg@zju.edu.cn}

\author{Zhaojing Luo}
\affiliation{%
  \department{School of Computing}
  \institution{National University of Singapore}
}
\email{zhaojing@comp.nus.edu.sg}

\author{Beng Chin Ooi}
\affiliation{%
  \department{School of Computing}
  \institution{National University of Singapore}
}
\email{ooibc@comp.nus.edu.sg}

\renewcommand{\shortauthors}{}

\begin{abstract}
Alphas are stock prediction models capturing trading signals in a stock market. A set of effective alphas can generate weakly correlated high returns to diversify the risk. Existing alphas can be categorized into two classes: Formulaic alphas are simple algebraic expressions of scalar features, and thus can generalize well and be mined into a weakly correlated set. Machine learning alphas are data-driven models over vector and matrix features. They are more predictive than formulaic alphas, but are too complex to mine into a weakly correlated set. In this paper, we introduce a new class of alphas to model scalar, vector, and matrix features which possess the strengths of these two existing classes. The new alphas predict returns with high accuracy and can be mined into a weakly correlated set. In addition, we propose a novel alpha mining framework based on AutoML, called AlphaEvolve, to generate the new alphas. To this end, we first propose operators for generating the new alphas and selectively injecting relational domain knowledge to model the relations between stocks. We then accelerate the alpha mining by proposing a pruning technique for redundant alphas. Experiments show that AlphaEvolve can evolve initial alphas into the new alphas with high returns and weak correlations.
\end{abstract}

\settopmatter{printacmref=false}
\maketitle
\begin{figure}[h]
\centering
\resizebox{\columnwidth}{!}{\includegraphics{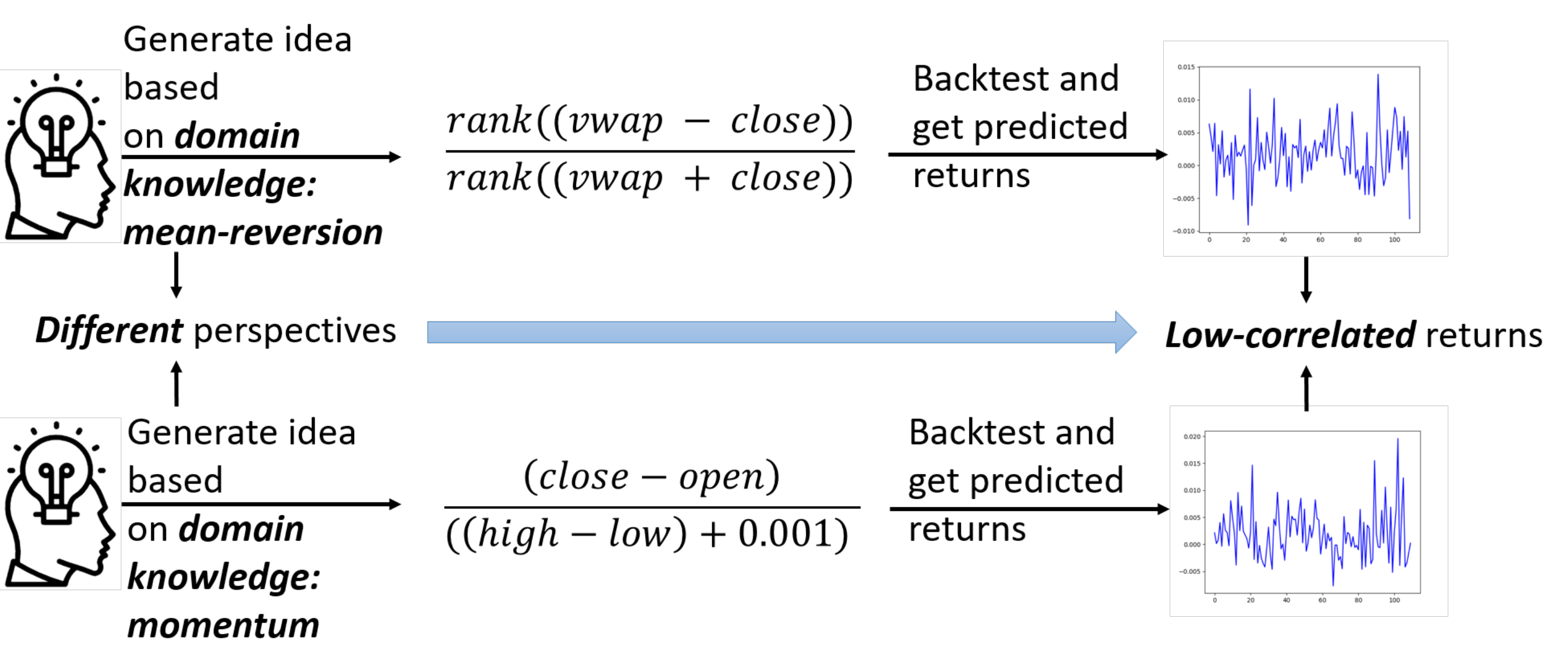}}
\caption{
The generating process of formulaic alphas.
}
\label{fig:uncorrelatedalpha}
\end{figure}
\section{Introduction}
\label{sec:Introduction}
Alphas are stock prediction models generating trading signals (i.e., triggers to buy or sell stocks). {Mining alphas with high returns}\footnote{In this paper, the return is defined as Today’s Stock Return = (Today’s Stock Price - Yesterday’s Stock Price) / (Yesterday’s Stock Price)} {has been an active research topic in the field of data mining \cite{10.1007/3-540-47887-6_48, zhang:SPP3097983.3098117, 10.1145/3308560.3317701, 10.1145/3318464.3389720}.} 
However, high returns typically come with high risks. 
To buffer risk, experts from hedge funds\footnote{Hedge funds are institutional investors and among the most critical players in the stock markets \cite{cao2018hedge}.} 
aim at identifying a set of weakly correlated\footnote{In this paper, we adopt the standard for weak correlation in hedge funds: the sample Pearson correlation of 15\% between portfolio returns of different alphas \cite{RePEc:arx:papers:1601.00991}.} formulaic alphas with high returns.
The identifying process is illustrated in Figure~\ref{fig:uncorrelatedalpha}. 
Alphas are first designed based on experts’ different perspectives and then backtested on the markets to ensure weakly correlated returns.
Such alphas, as algebraic expressions, have fine properties of simplicity and good generalizability \cite{cranmer2020discovering}. 

\begin{figure*}[t]
\centering    
  \includegraphics[width=\textwidth]{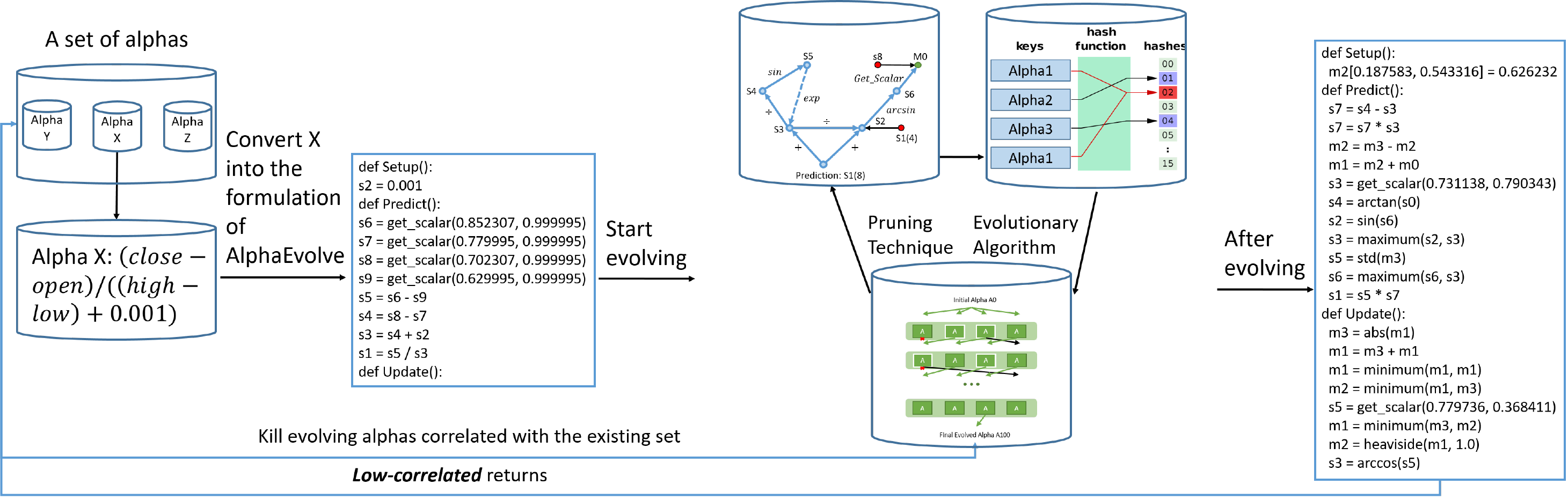}
  \caption{\textcolor{black}{AlphaEvolve architecture overview.}}
\label{Idea_Example}  
\end{figure*}
Existing AI approaches surpass the experts
by designing complex machine learning alphas
\cite{feng2019temporal,ding-etal-2016-knowledge, 8622541} or using the genetic algorithm to automatically mine a set of formulaic alphas \cite{huatai,huataiagain}.
In the first approach, machine learning alphas are machine learning models that generate trading signals.
They can model high-dimensional features and learn from training data to boost generalization performance. However, they are complex in their structures, and thus difficult to mine into a set of alphas with weakly correlated returns and to maintain this set \cite{luo2021mlcask}.
Further, machine learning alphas designed with domain knowledge are based on strong structural assumptions. 
To be specific, the injection of relational domain knowledge assumes the returns of the stocks from the same sector change similarly \cite{feng2019temporal}. 
However, this assumption is less likely to hold for a volatile stock market. In the second approach, the genetic algorithm has two limitations:
First, the search space of the algorithm is small because only arithmetic operations are considered.
In such a small search space, the algorithm can hardly evolve the initial alphas to improve the performance;
Second, the algorithm searches formulaic alphas, which only utilize short-term features (e.g., the moving average of last 30 days' close prices).

A framework called AutoML-Zero \cite{real2020automl} was recently proposed to discover a neural network from scratch and can thus be considered for alpha mining. Such consideration is advantageous over the genetic algorithm: AutoML-Zero expands the search space by allowing more operation types, i.e., vector or matrix operations, for feature capturing. 
Consequently, a domain-expert-designed alpha can be further improved. However, merely applying AutoML-zero to alpha mining suffers from the drawbacks of machine learning alphas, i.e., complex alpha structures and ineffective use of relational domain knowledge, leading to ineffective and inefficient alpha mining. Specifically, machine learning alphas typically contain complex components that are difficult to discover from scratch, e.g., graph neural network \cite{feng2019temporal}, attention mechanism and LSTM neural networks \cite{8622541}. Even discovering a two-layer neural network requires intensive computing resources and long searching time (i.e., 100 to 1000 processes over a week) to evaluate $10^{12}$ candidate alphas \cite{real2020automl}. Also, AutoML-Zero considers stocks as independent tasks and thus cannot leverage relational domain knowledge, which provides rich information for modeling the relations between stocks.

\begin{figure*}[h]
\centering    
  \includegraphics[width=\textwidth]{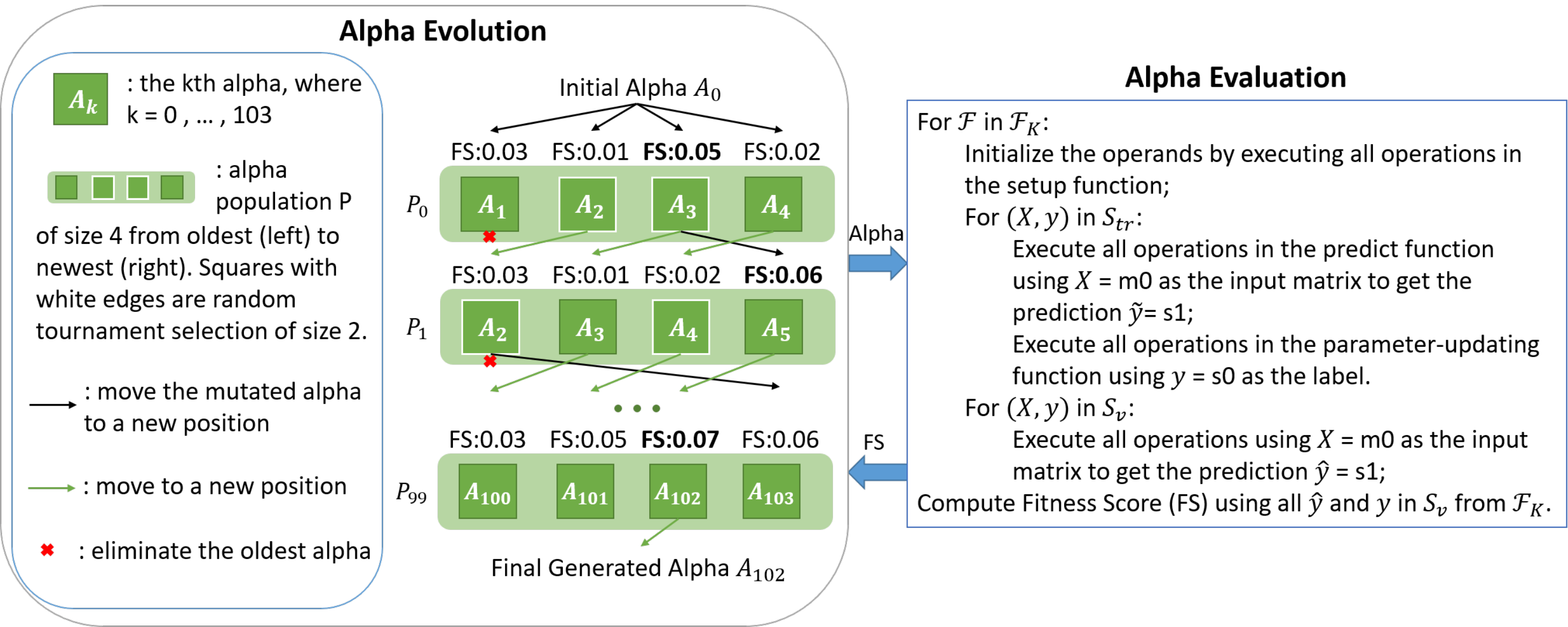}
  \caption{\textcolor{black}{The evolution and evaluation processes in alpha mining.}}
  \label{all_process}
\end{figure*}
To address these issues, we first introduce a new class of alphas, which can be effectively mined into a weakly correlated set of alphas. Next, we propose a novel alpha mining framework, AlphaEvolve, for generating the new alphas.
An overview of AlphaEvolve is illustrated in Figure~\ref{Idea_Example}. 
A well-designed formulaic alpha is first transformed into the formulation of AlphaEvolve. 
Then an evolutionary algorithm is initialized with this alpha. During the evolutionary process, a population of candidate alphas is iteratively updated for generating better alphas. Besides, candidate alphas are eliminated if they are correlated with a given set of alphas. 
Finally, an alpha is generated with weakly correlated high returns. 
To this end, we first introduce operators extracting scalar inputs from the input matrix. 
{These operators facilitate searching for the new alphas and avoid discovering a complex machine learning alpha from scratch, because the new alphas augmented with extracted inputs are more predictive and thus more likely to be selected in the evolutionary process than complex machine learning alphas, which are less predictive}. Next, we further introduce operators to selectively inject sector-industry relations of all stocks into an alpha. Lastly, we propose a pruning technique to boost search efficiency.
This technique helps to avoid redundant calculations in alpha search with a large number of tasks.

In this paper, we make the following contributions:
\begin{itemize}
  \item We first introduce a new class of alphas with intriguing strengths: like formulaic alphas, these alphas can model scalar features and thus are simple to mine into a weakly correlated set, but, like machine learning alphas, they are high-dimensional data-driven models utilizing long-term features.
  We then propose a novel alpha mining framework, AlphaEvolve, to generate the new alphas. 
  To the best of our knowledge, we are the first to solve the stock prediction problem based on AutoML and the first to tackle the problem of mining weakly correlated alphas. 
  \item We enable AlphaEvolve to selectively inject relational domain knowledge without any strong structural assumption in an alpha.
  \item \textcolor{black}{We propose an optimization technique to accelerate alpha mining by pruning redundant alphas.}  
  \item We conduct  extensive experimental study on AlphaEvolve using the stock price data of NASDAQ. 
  The results show that AlphaEvolve generates alphas with weakly correlated high returns. 
\end{itemize}

The remainder of the paper is structured as follows. Section \ref{Problem Formulation} defines the problem formulation. Section \ref{Evolutionary algorithm For Alpha Mining} introduces the evolutionary algorithm AlphaEvolve is based on. Section \ref{AlphaEvolve} elaborates on the optimization of AlphaEvolve. Section \ref{Experimental Study} analyses the experiment results. Section \ref{sec2} reviews the background of the alpha mining problem. Section \ref{sec5} concludes this paper. 

\section{Problem Formulation}
\label{Problem Formulation}
We adopt the alpha definition used by traders in hedge funds \cite{RePEc:arx:papers:1601.00991}. An alpha is a combination of mathematical expressions, computer source code, and configuration parameters that can be used, in combination with historical data, to make predictions about future movements of stocks \cite{book}. 
Under this definition, an alpha is formulated in AlphaEvolve as a sequence of operations, each of which comprises an operator (i.e., OP), input operand(s), and an output operand. For each operand, we use $s$, $v$, $m$ to denote a scalar, a vector, and a matrix respectively.
Let us refer back to Figure~\ref{Idea_Example}. 
The fifth operation $s5=s6-s9$ in the Predict() component of the alpha X before evolving has $s5$ as the output operand, $s6$ and $s9$ as the input operands, and $-$ as the OP. 
Specifically, the operand $s5$, $s6$, and $s9$ are the sixth, seventh, and tenth scalar operand respectively. 
Special operands are predefined, e.g., the input feature matrix $m0$, the output label $s0$, and the prediction $s1$. 
Note that an operand can be overwritten and therefore only the last $s1$ in the predict function is the final prediction. {The allowable OPs are composed of our proposed OPs and basic mathematical operators for scalars, vectors, and matrices \cite{real2020automl}.} 

Each alpha consists of three components: a setup function to initialize operands, a predict function to generate a prediction, and a parameter-updating function to update parameters if any parameters are generated during the evolution. 
These three components correspond to def Setup(), def Predict(), and def Update() respectively in the examples in Figure \ref{Idea_Example}. 
In the prediction function, the new class of alphas improves the formulaic alphas by extracting features from vectors and matrices. 
Additionally, including scalars into operations makes the new alphas less complex than machine learning alphas, which only allow operations of high-dimensional vectors or matrices.
In the parameter-updating function, we define the operands that are updated in the training stage (i.e., features) and passed to the inference stage as parameters. 
Unlike intermediate calculation results which are only useful for a specific prediction, these operands, as features of long-term training data, improve the alpha’s inference ability. 
The new alphas with these parameters are advantageous over formulaic alphas which cannot utilize long-term historical data.

We aim to search for the best alpha from all possible alphas constrained by the maximum allowable number of operations for each component, the allowable OPs for each component, and the maximum allowable number of operands for scalars, vectors, and matrices.

An alpha is evaluated over a set of tasks $\mathcal{F}_{{K}}$, where $K$ is the number of tasks.
Each task is a regression task for a stock, mapping an input feature matrix $\mathcal{X} \in \mathbb{R}^{f \times w}$ to a scalar label of return ${y}$, where $f$ is the number of feature types and $w$ is the input time window in days. 
The pair of $\mathcal{X}$ and ${y}$ defines a sample. 
All samples $S$ are split into a training set ${S}_{tr}$, a validation set ${S}_{v}$, and a test set ${S}_{te}$.

\section{Evolutionary Algorithm For Alpha Mining}
\label{Evolutionary algorithm For Alpha Mining}
AlphaEvolve is based on the evolutionary algorithm, which is an iterative selection process for the best alpha under a time budget. The process is illustrated in Figure \ref{all_process}:

(1) In the first iteration, AlphaEvolve is initialized by a starting parent alpha, \textcolor{black}{e.g., ${A}_{0}$ in Figure \ref{all_process}}. 
A population ${P}_{0}$ is generated by mutating the parent alpha. 
Two types of mutations are performed on the parent alpha to generate a child alpha: (1) randomizing operands or OP(s) in all operations; (2) inserting a random operation or removing an operation at a random location of the alpha.

(2) Each alpha of ${P}_{0}$ is evaluated on the tasks $\mathcal{F}_{{K}}$. The evaluation process outputs a fitness score as shown on the right side of Figure \ref{all_process}. 
We use the Information Coefficient (IC) as the fitness score for alpha $i$ (Eq. \ref{equ:1}), where ${\mathbf{\hat{y}} }_{t}^{(i)}=\left({\hat{y}}_{t, 1}^{(i)}, \ldots, {\hat{y}}_{t, K}^{(i)}\right)$ is the vector of predictions at date $t$, ${\mathbf{{y}}}_{t}$ is the vector of corresponding labels, corr is the sample Pearson correlation, $N$ is the number of samples in ${S}_{v}$.
\begin{equation}
\setlength\abovedisplayskip{8pt}
\setlength\belowdisplayskip{8pt}
\textcolor{black}{I C_{i}=\frac{1}{N} \sum_{t=1}^{N} {corr}\left({\mathbf{\hat{y}}}_{t}^{(i)}, {\mathbf{{y}}}_{t}\right)}
\label{equ:1}
\end{equation}

{(3) The alpha with the highest fitness score in a randomly selected set of fixed size, called the tournament, is selected as the new parent alpha. For example, the alpha with the highest fitness score in the tournament of ${P}_{0}$, ${A}_{3}$, is selected as the parent alpha.} 

{(4) In the subsequent iterations, a new population is generated by adding the mutated parent alpha into the previous population and eliminating the oldest alpha.
For example, ${P}_{1}$ is generated by adding ${A}_{5}$ mutated from ${A}_{3}$ and eliminating ${A}_{1}$ from ${P}_{0}$ in Figure \ref{all_process}.} 

(5) If the training budget is exhausted, the alpha with the highest fitness score in the population is selected as the evolved alpha.
\begin{figure}[ht]
\centering
\resizebox{\columnwidth}{!}{\includegraphics{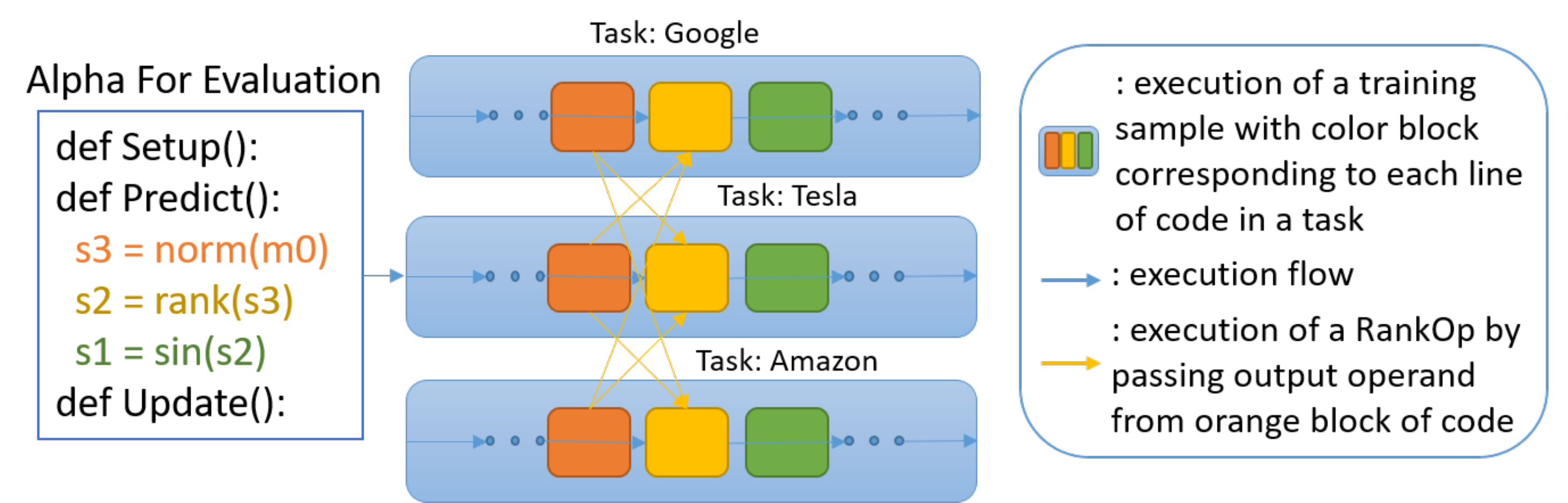}}
\caption{\textcolor{black}{An example of the execution of the RelationOp.}}
\label{fig:efficientExecute}
\end{figure} 
\section{Optimization}
\label{AlphaEvolve}
{This section introduces the proposed OPs and the pruning technique for redundant alphas to optimize the alpha search.}
\subsection{RelationOp and ExtractionOp}
\label{RelationOp}
Unlike conventional AutoML frameworks where tasks
are mutually independent \cite{article,real2020automl}, our tasks $\mathcal{F}_{{K}}$ are related because the stocks predicted in $\mathcal{F}_{{K}}$ are classified into sectors and industries in a stock market. 
Modeling such relations in a stock market effectively improves the prediction accuracy \cite{feng2019temporal}. 
{We design a set of OPs, called RelationOps, to model such relations: they calculate an output operand based on input scalar operands calculated in the current task and other related tasks in the same sector (industry).}

For the execution of an operation with the RelationOp on a sample ${s}^{(a)} \in S$ where $a \in \mathcal{F}_{{K}}$, the input operands are the output operands calculated on the samples $\mathbf{s}^{\mathcal{F}} \subset S$  where \textcolor{black}{$\mathcal{F}$ is a set of related tasks}. 
For the example shown in Figure \ref{fig:efficientExecute}, the execution of $s2 = rank(s3)$ (in yellow) has its inputs calculated as $s3 = norm(m0)$ (in orange) on the samples of the same time step from all related tasks, where $norm$ calculates the Frobenius norm of a matrix.
The output operand and \textcolor{black}{$\mathcal{F}$ are determined by the types of RelationOps}: (1) RankOp outputs the ranking of the input operand calculated on ${s}^{(a)}$ among those calculated on $\mathbf{s}^{\mathcal{F}_{{K}}}$; 
(2) RelationRankOp outputs the ranking of the input operand calculated on ${s}^{(a)}$ among those calculated on $\mathbf{s}^{\mathcal{F}_{{I}}}$ \textcolor{black}{where $\mathcal{F}_{{I}} \subset \mathcal{F}_{{K}}$ are the tasks in the same sector (industry)};
(3) RelationDemeanOp calculates the difference between the input operand calculated on ${s}^{(a)}$ and the mean of those calculated on $\mathbf{s}^{\mathcal{F}_{{I}}}$.

{We define OPs extracting a scalar feature from $\mathcal{X}$ and OPs extracting a vector feature from $\mathcal{X}$ as GetScalarOps and GetVectorOps respectively, or called ExtractionOps in general.} Once an ExtractionOp is selected in a mutation step,
$\mathcal{X}$ serves as a pool for selecting a scalar or a vector, and thus the actual input of an alpha can be $\mathcal{X}$ or just a scalar, a column, or a row of $\mathcal{X}$.

ExtractionOps facilitate searching for the new class of alphas and avoid discovering a complex machine learning alpha from scratch. 
Specifically, in the evolutionary process, the fitness score of an alpha augmented with extracted scalar inputs is usually high among a population, and thus this alpha is likely to survive to the next population as a parent alpha. 
Iteratively this process guides the evolution towards the new alpha instead of a machine learning alpha that does not allow scalar inputs.
\begin{figure}
\begin{subfigure}{.5\textwidth}
  \centering
  \includegraphics[width=\linewidth]{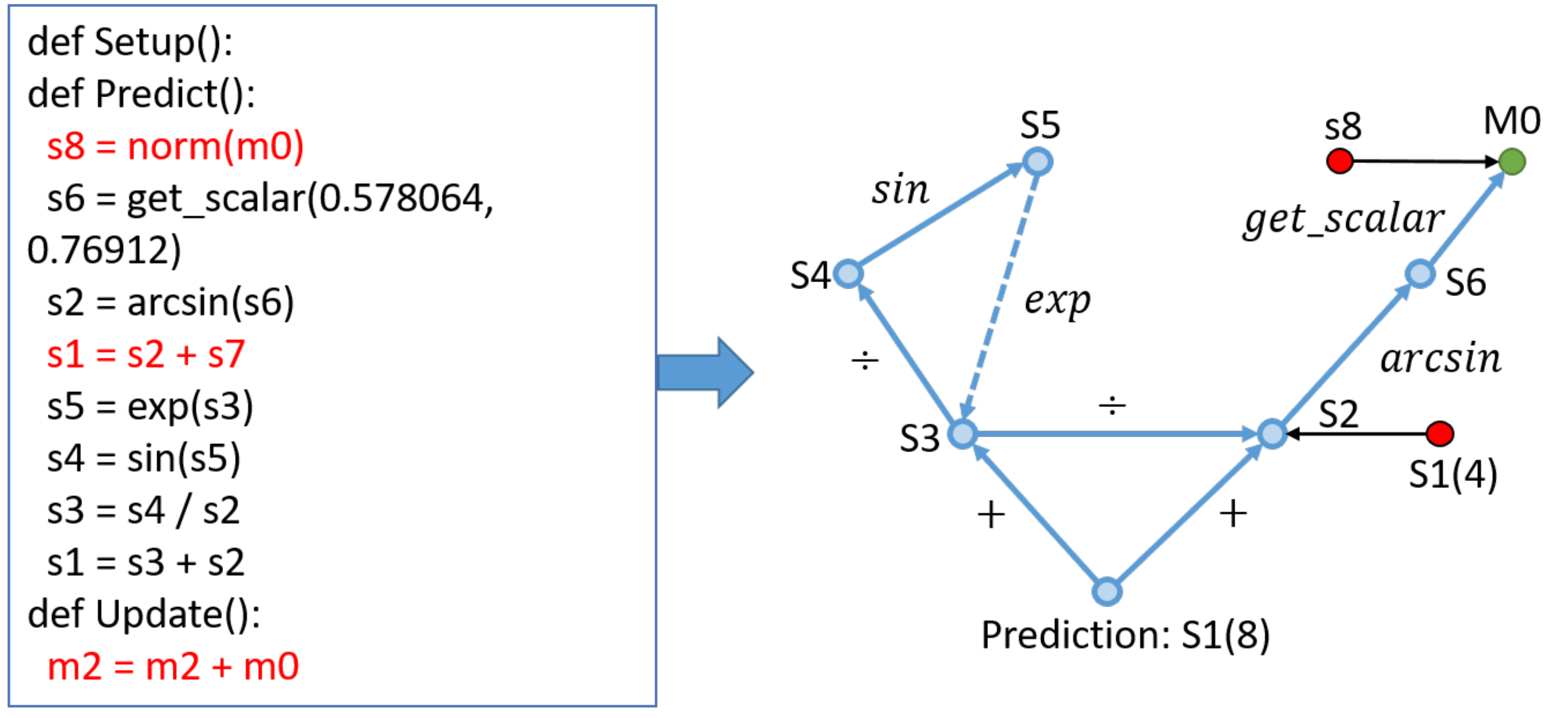}  
  \caption{{An alpha with redundant operations.}}
  \label{validity_check_a}
\end{subfigure}
\par\bigskip
\begin{subfigure}{.5\textwidth}
  \centering
  \includegraphics[width=\linewidth]{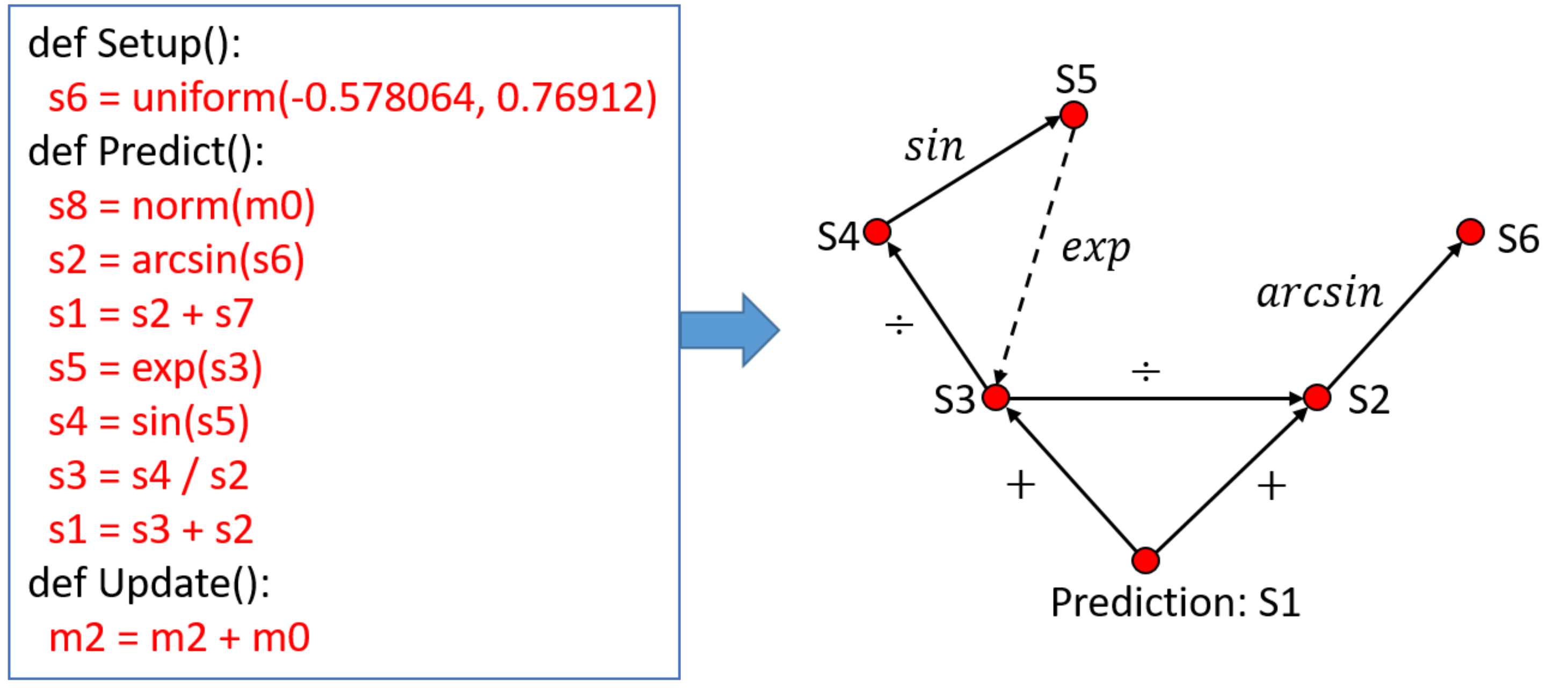}  
  \caption{{A redundant alpha.}}
  \label{validity_check_b}
\end{subfigure}
\caption{{Examples of the redundancy pruning process.}}
\label{validity_check}
\end{figure}
\subsection{\textcolor{black}{Pruning Technique}}
\label{Query Form of An Alpha}
Searching for the best alpha in a large space efficiently is challenging. 
Earlier work \cite{real2020automl} increases efficiency by avoiding repeated evaluations of an alpha. It fingerprints an alpha by its predictions on a small set of samples and uses the fingerprint to terminate any repeated alphas with the same predictions on the set of samples in subsequent searches. 
However, this method is inefficient for two reasons. First, this method evaluates redundant operations and alphas.
Second, the cost of testing on the set of samples is higher in our problem because the number of tasks $K$ is larger than the number of the original problem. 
Further, we cannot approximate all the tasks with a small subset because the stocks predicted in the tasks vary greatly given the noisy nature of the stock price data.

We thus propose an optimization technique by pruning redundant operations and alphas as well as fingerprinting without evaluation. 
Specifically, the fingerprint of an alpha is built by pruning redundant operations and alphas before evaluation, and transforming the strings of the alpha’s remaining operations into numbers. 
If the fingerprint is matched in the cache, the fitness score stored in the cache is reused. 
Otherwise, the alpha is evaluated to get its fitness score and then hashed into the cache.

The redundancy pruning process prunes the operations that do not contribute to the calculation between the input feature matrix $m0$ and the prediction $s1$. 
The process works as follows. 
First, we represent an alpha as a graph with operators as edges and operands as nodes. 
The prediction $s1$ is the root node. Next, starting from the root node,
we iteratively check a node’s redundancy by finding the operation where the node is the output operand and checking if the input operands of this operation are redundant nodes.
The check returns true if a leaf is $m0$. Finally, we prune the operation with a redundant output operand. 
This process is illustrated in Figure \ref{validity_check}, where the red, blue, and green nodes are the redundant operands, the necessary operands, and $m0$ respectively, and the dashed edge is the operator with the input operand calculated at the last time step.
In Figure~\ref{validity_check_a}, part of the alpha are redundant operations: 
The fourth operation of the Predict() component with output $s1$, denoted by $s1(4)$ in the graph, is redundant since it is overwritten by $s1(8)$, which is used as the prediction; The operation with output $s8$ is redundant since $s8$ does not contribute to the calculation of $s1(8)$.
In Figure~\ref{validity_check_b},
the alpha is redundant since $m0$ is not used to calculate the prediction.

We discuss two typical scenarios in the evolutionary process on how the pruning technique increases search efficiency.
In the early stage of the evolutionary process, an alpha usually has more redundant operations than useful ones.
These redundant operations can be pruned by the pruning technique.
In the later stage, an alpha with no redundancy tends to be vulnerable to random mutations, e.g., deleting a random operation would invalidate the prediction.
Consequently, this alpha would possibly become a redundant alpha after random mutations and would be pruned. 

\section{Experimental Study}
\label{Experimental Study}
In this section, we compare the performance of AlphaEvolve among different initializations, and with baselines including the genetic algorithm and complex machine learning alphas. Then we study the effectiveness of the parameter-updating function, the selective injection of relational domain knowledge, and the pruning technique.
\subsection{Dataset}
\label{Dataset}
We use the 5-year (2013-2017) stock price data from a major stock market NASDAQ. 
The 5-year data consists of 1220 days in total and is split into sets of 988, 116, and 116 days for training, validation, and test, respectively. 
Two types of stocks are filtered out in the data preprocessing stage: (1) the stocks without sufficient samples and (2) the stocks reaching too low prices during the selected period. The first is because they are less traded and thus only bring the noise to the model, while the second is because they are too risky for investors.
After filtering, there are $1026$ stocks left.
Each type of the features is normalized by its maximum value across all time steps for each stock. 

\subsection{\textcolor{black}{Baselines and Settings}}
\label{Comparison Algorithms}
We use the following baselines for comparison:
\begin{enumerate}
    \item {$alpha\_G$ is the searched alpha by the genetic algorithm, which is a popular alpha mining approach discussed in Section \ref{sec:Introduction}}.
    \item Rank\_LSTM is a variant of the LSTM model with its output mapped to a fully connected layer.
    \item RSR is a variant of Rank\_LSTM by adding a graph component, which is designed with the injection of relational domain knowledge by connecting stocks in the same sector (industry). The RSR model is reported with the best performance in the dataset \cite{feng2019temporal}.
    \item $alpha\_AE\_D$ is the evolved alpha by AlphaEvolve, initialized with a domain-expert-designed alpha. See details of the domain-expert-designed alpha in the alpha before evolving in Figure \ref{Idea_Example}.
    \item $alpha\_AE\_NOOP$ is the evolved alpha by AlphaEvolve with no initialization.
    \item $alpha\_AE\_R$ is the evolved alpha by AlphaEvolve, initialized with an alpha designed randomly.
    \item $alpha\_AE\_NN$ is the evolved alpha by AlphaEvolve, initialized with a two-layer neural network alpha.
\end{enumerate}

We shall now describe the setting for each method. For AlphaEvolve, we use the population size of 100 and the tournament size of 10. The mutation probability of each operation is set to 0.9.
The dimensions $f$ and $w$ for the input feature matrix $\mathcal{X}$ are 13.
The first four features are the moving averages of the close prices over 5, 10, 20, and 30 days, respectively. The next four are the close prices’ volatilities over 5, 10, 20, and 30 days, respectively. The last five are the open price, the high price, the low price, the close price, and the volume, respectively. The minimum number of the operations in each function is set to 1 and the maximum number to 21, 21, and 45, respectively. 
We choose the size of the maximum allowed scalar, vector, and matrix operands to be 10, 16, and 4, respectively. 
During the evolutionary process, we train our alpha by one epoch for fast evaluation. 

For the genetic algorithm, the input and the output are the same as those of AlphaEvolve. The sizes of the generation and the tournament are the same as AlphaEvolve. 
Apart from that, we follow the implementation details in \cite{huatai}: the probability of crossover, subtree mutation, hoist mutation, point mutation, and point replace are set to 0.4, 0.01, 0, 0.01, and 0.4, respectively. 

For Rank\_LSTM and RSR, each model’s input is a vector of the close prices’ moving averages over 5, 10, 20, and 30 days for each of the input stocks, while the output is the predicted return.
Following the experiment settings of \cite{feng2019temporal}, we fine-tune the hyper-parameters for Rank\_LSTM. To be specific, the grid for the sequence length, the number of units, and the hyperparameter balancing the loss terms are [4, 8, 16, 32], [32, 64, 128, 256], and [0.01, 0.1, 1, 10], respectively.
The learning rate is set to 0.001.
The best set of the hyperparameters is selected based on the performance on ${S}_{v}$.
This set is used for reporting the performance of Rank\_LSTM and then getting the pre-trained embeddings for RSR following the original implementation. The average of the testing results is reported by performing 5 runs with different random seeds.

\subsection{Evaluation Metrics}
\label{Evaluation Metric}
Apart from the IC, we use the Sharpe ratio to measure risk-adjusted returns of a portfolio built based on an alpha. 
We first introduce how this portfolio is built. After that, we define the portfolio return and the Sharpe ratio.

We use the long-short trading strategy, a popular hedge fund strategy \cite{RePEc:arx:papers:1601.00991}, to build a portfolio to evaluate an alpha’s investment performance. {At a time step $t$, the strategy selects stocks based on the ranking of predicted returns of all stocks to long (i.e., buy) and to short (i.e., borrow stocks to sell). The long position $V_{l}^{t}$ is built by buying the stocks with the top 50 predicted returns. 
The long position gains when the prices go up and the stocks are sold to win the price differences. The short position $V_{s}^{t}$ is built by borrowing the stocks with the bottom 50 predicted returns and selling them for cash.} The short position gains when the prices go down and the stocks are bought and returned to win the price differences. These long-short positions are balanced by the cash position $C^{t}$.
The reason is that with $V_{l}^{t}$ and $V_{s}^{t}$ fluctuating with gains or losses, we want to stick to a fixed investment plan (i.e., a fixed ratio between the two positions) to avoid large risk exposure on either side. 
The net asset value (NAV) is defined as $NAV^{t} = V_{l}^{t} + V_{s}^{t} - C^{t}$ and the portfolio return is defined as $R_{p}^{t} = (NAV^{t} - NAV^{t - 1})/NAV^{t - 1}$. 
The portfolio returns on $S_{v}$ or $S_{te}$ is a vector $\mathbf{R}_{p}=\left(R_{p}^{1}, \ldots, R_{p}^{N}\right)$, where $N$ is the size of $S_{v}$ or $S_{te}$.

The Sharpe ratio is defined as $SR = (\bar{R}_{p} - R_{r})/\sigma_{p}$, where $\bar{R}_{p}$ is the average of $\mathbf{R}_{p}$, $R_{r}$ is the risk-free rate\footnote{Following \cite{RePEc:arx:papers:1601.00991}, we set $R_{r}$ equal to 0 for simplicity.}, and $\sigma_{p}$ is the volatility of the portfolio calculated as the standard deviation of $\mathbf{R}_{p}$. Both $\bar{R}_{p}$ and $\sigma_{p}$ are annualized over 252 trading days. 

\subsection{Performance Evaluation}
\label{Performance Study}
\begin{figure*}[t]
\begin{subfigure}{0.18\textwidth}
\centering
\includegraphics[width=\textwidth]{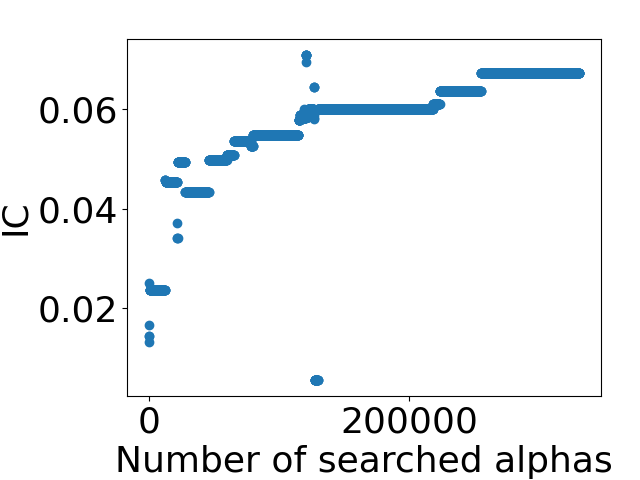}%
\caption{$alpha\_AE\_D\_0$}
\label{fig:trajectory for best alpha in round 0}
\end{subfigure}\hspace{\fill}
\begin{subfigure}{0.18\textwidth}
\centering
\includegraphics[width=\textwidth]{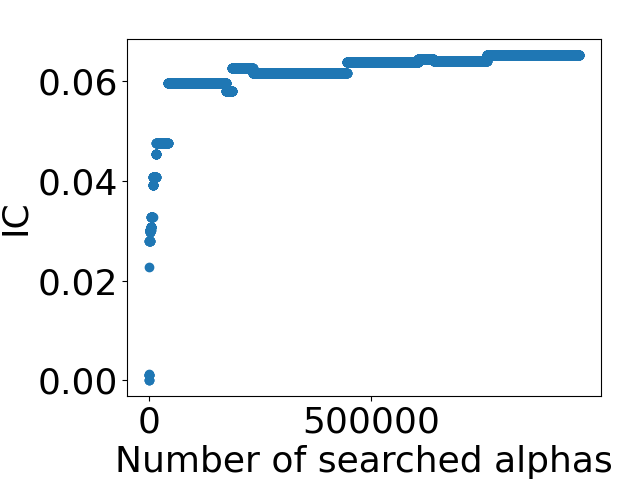}%
\caption{$alpha\_AE\_NN\_1$}
\label{fig:trajectory for best alpha in round 1}
\end{subfigure}\hspace{\fill}
\begin{subfigure}{0.18\textwidth}
\centering
\includegraphics[width=\textwidth]{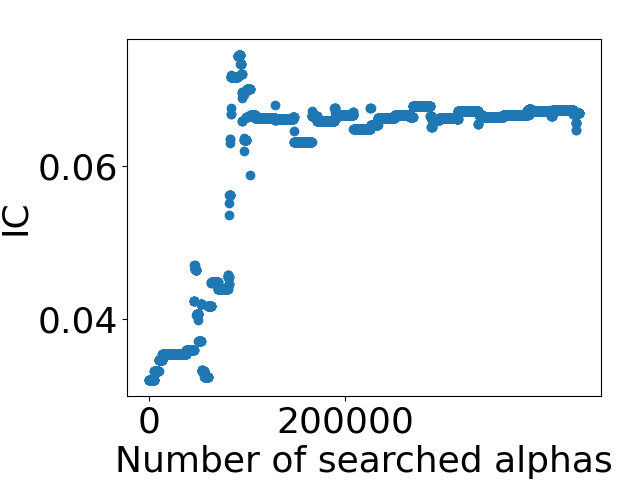}%
\caption{$alpha\_AE\_R\_2$}
\label{fig:trajectory for best alpha in round 2}
\end{subfigure}\hspace{\fill}
\begin{subfigure}{0.18\textwidth}
\centering
\includegraphics[width=\textwidth]{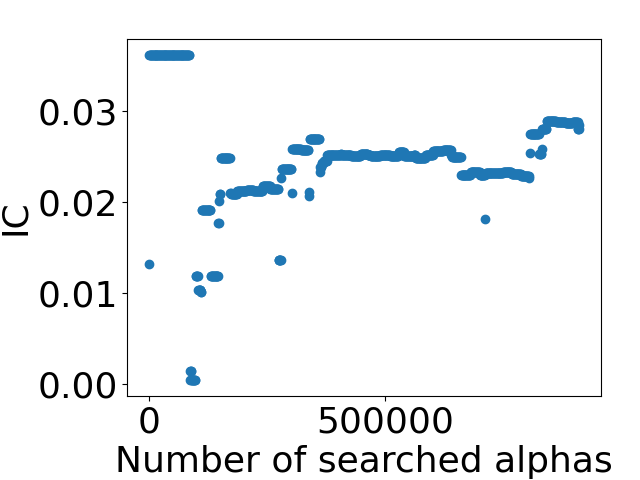}%
\caption{$alpha\_AE\_D1\_3$}
\label{fig:trajectory for best alpha in round 3}
\end{subfigure}\hspace{\fill}
\begin{subfigure}{0.18\textwidth}
\centering
\includegraphics[width=\textwidth]{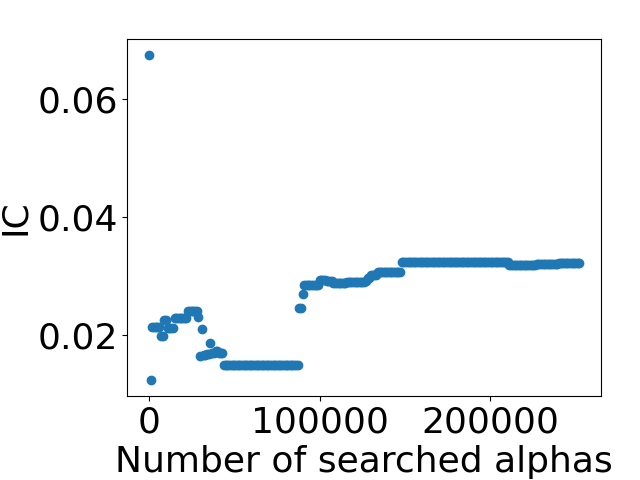}%
\caption{$alpha\_AE\_B0\_4$}
\label{fig:trajectory for best alpha in round 4}
\end{subfigure}\hspace{\fill}
\caption{\textcolor{black}{Evolutionary trajectories for the best alphas in all rounds.}}
\label{fig:trajectory for best alphas in all rounds}
\end{figure*}

\subsubsection{Alpha Mining Performance Comparisons}
\label{Alpha Mining Performance Comparison (Q1 and Q2)}
\label{Alpha Mining Performance Comparison}
\begin{table}[t]
\captionof{table}{Mining weakly correlated alpha with an existing domain-expert-designed alpha.}
\label{tab:Mining weakly correlated Alpha With an Existing domain-expert-designed Alpha}
\centering
\resizebox{\columnwidth}{!}{\begin{tabular}{lrrr}  
\hline
\textbf{Alpha}  & \textbf{Sharpe ratio} & \textbf{IC} & \textbf{Correlation with the existing alpha} \\
\hline
 alpha\_D\_0   &   4.111784&0.013159&NA\\
 alpha\_AE\_D\_0   &  \textbf{21.323797}&\textbf{0.067358}&0.030301
\\
 alpha\_G\_0  &13.034052&0.048853&-0.103120
\\
\hline
\end{tabular}}
\end{table}

We run AlphaEvolve with each of the initializations and the genetic algorithm for five rounds. {For AlphaEvolve, the best alpha with the highest Sharpe ratio among all initializations is selected into a set $\mathcal{A}$ after each round.} To achieve low correlation, we discard alphas correlated with any alpha in $\mathcal{A}$ above a cutoff during the evolutionary process, where the correlation is calculated using portfolio returns on ${S}_{v}$ and the cutoff is set at 15\% by the standard in hedge funds \cite{RePEc:arx:papers:1601.00991}. 
Note that as the number of rounds and the size of $\mathcal{A}$ increase, the difficulty of generating an alpha with low correlation increases. The same process is applied to the genetic algorithm. The time budget is set to 60 hours in each round. 
Notation-wise, we use the last digit number in the alpha name to represent the round number (starting from 0). 
In the last round, the alphas in $\mathcal{A}$ are used as initializations, each denoted with $B$ followed by a number referring to the round generating the alpha. Note that we do not set a cutoff with the domain-expert-designed alpha $alpha\_D\_0$ because its Sharpe ratio and IC are too low compared to the evolved alphas and its correlations with the evolved alphas are weak.
These results are observed in Table \ref{tab:Mining weakly correlated Alpha With an Existing domain-expert-designed Alpha}, in which $alpha\_AE\_D\_0$ and $alpha\_G\_0$, with the cutoffs set for the correlations with $alpha\_D\_0$, are compared with $alpha\_D\_0$.

The comparison results with the genetic algorithm are shown in Table \ref{tab:compare_all_alphas_Q1}, where the best alpha in a round is marked in bold. 
We observe that the Sharpe ratio and the IC of the genetic algorithm deteriorate with more cutoffs, showing that it does not do well in mining weakly correlated alphas. 
This result is due to the smaller search space of the genetic algorithm. 
Due to the consecutive low performances of $alpha\_G\_2$ and $alpha\_G\_3$, we stop the search for $alpha\_G\_4$.
\begin{table}[t]
\captionof{table}{{Performance of weakly correlated alpha mining.}}
\label{tab:compare_all_alphas_Q1}
\centering
\resizebox{\columnwidth}{!}{\begin{tabular}{ lrrr }  
\hline
 \textbf{Alpha}  & \textbf{Sharpe ratio} & \textbf{IC} & \textbf{Correlation with the best alphas} \\
\hline
 alpha\_AE\_D\_0   &   \textbf{21.323797}&\textbf{0.067358}&NA\\
 alpha\_G\_0  &13.034052&0.048853&NA\\
 \hline
 alpha\_AE\_D\_1   & \textbf{13.580572}&\textbf{0.056703}&-0.303845\\
 alpha\_G\_1  &4.407823&0.037521&-0.267428\\
 \hline
 alpha\_AE\_D\_2   &\textbf{15.067808}&\textbf{0.052464}&-0.040815\\
 alpha\_G\_2  &-1.936161&0.000779&-0.065011\\
 \hline
 alpha\_AE\_D\_3   &\textbf{4.901069}&\textbf{0.028437}&-0.202224\\
 alpha\_G\_3  &-1.971355&0.000000&-0.014054\\
 \hline
 alpha\_AE\_B0\_4 &\textbf{9.502871}&\textbf{0.032155}&0.137851\\
 alpha\_G\_4  &NA&NA&NA\\
\hline
\end{tabular}}
\end{table}

\begin{table}[t]
\captionof{table}{{Performance of weakly correlated alpha mining for different initializations.}}
\label{tab:compare_all_alphas_Q2}
\centering
\resizebox{\columnwidth}{!}{\begin{tabular}{ lrrr }  
\hline
\textbf{Alpha}  & \textbf{Sharpe ratio} & \textbf{IC} & \textbf{Correlation with the best alphas} \\
\hline
 alpha\_AE\_D\_0   &   \textbf{21.323797}&\textbf{0.067358}&NA\\
 alpha\_AE\_NOOP\_0   &  12.126316&0.046382&NA\\
 alpha\_AE\_R\_0   &10.718915&0.047780&NA\\
 alpha\_AE\_NN\_0   & 14.576585&0.057008&NA\\
 \hline
 alpha\_AE\_D\_1   & 13.580572&0.056703&-0.303845\\
 alpha\_AE\_NOOP\_1   &11.858020
&0.044230&-0.393348\\
 alpha\_AE\_R\_1   &12.186153&0.050688&0.008778\\
 alpha\_AE\_NN\_1   & \textbf{14.175835}&\textbf{0.065209}&-0.240786\\
 \hline
 alpha\_AE\_D\_2   &15.067808&0.052464&-0.040815\\
 alpha\_AE\_NOOP\_2   &12.309789&0.051791&-0.337799\\
 alpha\_AE\_R\_2   &\textbf{18.629571}&\textbf{0.066962}&-0.177144\\
 alpha\_AE\_NN\_2   &13.606091&0.052847&-0.242090\\
 \hline
 alpha\_AE\_D\_3   &\textbf{4.901069}&0.028437&-0.202224\\
 alpha\_AE\_NOOP\_3   &3.873076&0.015012&0.007269\\
 alpha\_AE\_R\_3   &3.660071&0.023426&-0.002207\\
 alpha\_AE\_NN\_3   &3.070874&\textbf{0.031879}&-0.033980\\
 \hline
 alpha\_AE\_B0\_4 &\textbf{9.502871}&\textbf{0.032155}&0.137851\\
 alpha\_AE\_B1\_4   &3.649546&0.014045&-0.003357\\
 alpha\_AE\_B2\_4   &12.275912&0.059586&0.217138\\
 alpha\_AE\_B3\_4
   &3.803120&0.021081&-0.064233\\
\hline
\end{tabular}}
\end{table}

The comparison results among different initializations are shown in Table \ref{tab:compare_all_alphas_Q2}.
$alpha\_AE\_D\_x$ shows the best results for $x = 0,3$ as well as $alpha\_AE\_B0\_4$ in the last round. These show that AlphaEvolve can generate good alphas by leveraging a well-designed alpha. The decrease in performance for $alpha\_AE\_D\_1$ and $alpha\_AE\_D\_2$ is because the cutoff is set for the correlations with $alpha\_AE\_D\_0$, which has the same initialized alpha. For the same reason, after the cutoff is set for the correlations with $alpha\_AE\_NN\_1$, $alpha\_AE\_NN\_x$ has dropped in performance significantly for $x = 2,3$. 
A similar drop in performance is also observed for $alpha\_AE\_$ $R\_3$. 
Such drops are observed for most initializations, but $alpha\_AE\_$ $D\_x$ is nonetheless the second-best for $x = 1,2$. $alpha\_AE\_NOOP$ $\_x$ shows the worst performances by not being the best alpha in any round due to no initialization.

The first four rounds show a decreasing trend in both the Sharpe ratio and the IC because the accumulative cutoffs increase the search difficulty. This increasing difficulty is also shown in the evolutionary trajectories of the best alphas from all rounds in $\mathcal{A}$ in Figure \ref{fig:trajectory for best alphas in all rounds}, where the ICs decrease and fluctuate as the round increases. This trend reverses when we set the previous best alphas as the initialized alphas in the last round. It shows the capability of AlphaEvolve in generating weakly correlated alphas with all cutoffs. 
Note that we choose $alpha\_AE\_B0\_4$ as the best alpha instead of $alpha\_AE\_B2\_4$ since the latter correlates with the previous alphas above 15\%. 
\subsubsection{Study of The Evolved Alphas}
\label{Study of Evolved Alphas (Q4)}
 We first study the best alphas from all rounds in $\mathcal{A}$ and then perform the ablation study on the parameter-updating functions. 
 To ease readability, we first change the raw output of AlphaEvolve, e.g., the form of the evolved alpha in Figure \ref{Idea_Example}, into a compact set of equations.
 Then we divide the equations into three parts: $M$, $P$, and $U$.
 In the training stage, the predict function is $M$ and the parameter-updating function is $U$, while in the inference stage, the prediction function consists of $M$ and $P$.
 $M$ is used in both stages to pass parameters between the stages.

For $alpha\_AE\_D\_0$ (Eq. \ref{equ:2} to Eq. \ref{equ:9}), $S4_{t-2}$ and $S2_{t-2}$ are updated by Eq. \ref{equ:6} and Eq. \ref{equ:7} respectively in the training stage, and then passed to $M$ as the 
parameters at the beginning of the inference stage.
These parameters affect output operands $S1_{t}$ in Eq. 2 and $S3_{t-1}$ in Eq. 3 by initializing the input operands.
Then the parameter $S4_{t-2}$ is overwritten in Eq. \ref{equ:4}. 
The remaining parameter $S2_{t-2}$ is used in an upper bound $\arcsin(S 2_{t-2})$ for an expression of the temporal difference (i.e., trend) of the high prices in Eq. \ref{equ:3}.
This bound will be overwritten once it is less than the trend, upon which the model becomes a formulaic alpha. {This is because a formulaic alpha is a special case of the new alpha with no parameters.} The prediction is the fraction with an expression of bounded trend feature on high prices as the numerator and another trend feature as the denominator in Eq. \ref{equ:2}. Therefore, this alpha makes trading decisions based on the trends of high prices, constrained by the historically updated bound $\arcsin(S 2_{t-2})$.
\small
\begin{align}
M:\quad S 1_{t}  &={\tan (S 3_{t-1})}/{\cos (S4_{t-2}-\arcsin ({high\_price}_{t-1}))} \label{equ:2} \\ 
S 3_{t-1}&=\min (S 4_{t-2}-\arcsin({high\_price}_{t-1}),\arcsin(S 2_{t-2})) \label{equ:3} \\ 
P:\; S 4_{t-2}&=\arctan (\arcsin ({high\_price}_{t-3})) &\label{equ:4} \\ 
S 2_{t-2}&=\arctan (S 3_{t-2})& \label{equ:5} \\
U:\, S 4_{t-2}&=\tan ({heaviside}(S 1_{t-2})) &\label{equ:6} \\
S 2_{t-2}&=\arccos (n o r m ( {norm}(M 2_{t-4}, axis=0))) &\label{equ:7} \\
M 2_{t-4}&=\min ({abs} ({abs} (M 1_{t-4})), \nonumber \\
&{broadcast}({broadcast}(S0_{t-4}),axis=1)) &\label{equ:8} \\
M 1_{t-4}&={matmul}(M2_{t-5},M1_{t-5}) \label{equ:9}
\end{align}

\normalsize
For $alpha\_AE\_NN\_1$ (Eq. \ref{equ:10}), it is a complex formula using $relation\_rank$ operator and $high\_price$.
\small
\begin{align}
M: S1_{t}=& \log (\cos (\arcsin (\min ({tsrank}({abs}( {relation\_rank} \nonumber \\
& (\arctan (\sin (\sin (\exp ( {high\_price}_{t-2}))))))))), \nonumber \\
& \log (\sin (\arctan (\sin (\sin (\exp ({high\_price}_{t-1})))))))) \label{equ:10}
\end{align}
\normalsize
For $alpha\_AE\_R\_2$ (Eq. \ref{equ:11} to Eq. \ref{equ:16}), the parameter $M 2_{t-2}$ is updated recursively with an expression of the input feature matrix $M0_{t-2}$ (Eq. \ref{equ:14} and Eq. \ref{equ:15}). $S 2_{t-2}$ is a trend feature based on the comparison between ${ high\_price }_{t-4}$ and a recursively compared feature of ${ high\_price }_{t-5}$ (Eq. \ref{equ:12} and Eq. \ref{equ:13}). Thus in the inference stage, we can observe from Eq. \ref{equ:11} that the alpha makes trading decision based on the volatility of the historically updated features $M 2_{t-2}$, the trend feature based on high prices $S 2_{t-2}$ and the recent return $S 0_{t-3}$. 
\small
\begin{align}
M:\quad \; S1_{t}&= std\left(M 2_{t-2}\right) \times \left(\arctan \left(S 0_{t-3}\right)-S 2_{t-2}\right) \times S 2_{t-2} \label{equ:11} \\
P:\ \; \, S2_{t-2}&= \max \left(\sin \left(S 3_{t-3}\right), { high\_price }_{t-4}\right) \label{equ:12} \\
S3_{t-3}&= \max \left(S 3_{t-4}, \max \left(\sin \left(S 3_{t-4}\right), {high\_price }_{t-5}\right)\right) \label{equ:13} \\
U:\, M 2_{t-2} &= a b s\left(M 1_{t-2}\right) \label{equ:14} \\
M 1_{t-2} &= M 2_{t-3}+{ heaviside }(\min (M 2_{t-3}, \nonumber \\
& \min (M 2_{t-3}+M 1_{t-3}, M 2_{t-3})), 1)+M 0_{t-2} \label{equ:15} \\
S 2_{t-2} &= {low\_price}_{t-10} \label{equ:16}
\end{align}
\normalsize
For $alpha\_AE\_D\_3$ (Eq. \ref{equ:17} to Eq. \ref{equ:19}), a lower bound of the transpose of the input feature matrix $M 0_{t-2}$ is set for an expression of $M 0_{t-2}$ to recursively update the parameter $M1_{t-2}$ (Eq. \ref{equ:19}). At the beginning of the inference stage, $M1_{t-2}$ is passed to $M$ and $P$ as initial matrices $M1_{t-2}$ and $M1_{t-3}$ respectively (Eq. \ref{equ:17} and Eq. \ref{equ:18}). Then $M1_{t-3}$ recursively compares with an expression of $M 0_{t-2}$ (Eq. \ref{equ:18}). Finally, the prediction is the standard deviation of another comparison result between $M1_{t-2}$ and an expression of another input feature matrix $M0_{t-1}$ (Eq. \ref{equ:17}), showing that this alpha trades based on the volatility of an expression of the previous day’s features $M 0_{t-1}$ bounded by the historically updated features $M1_{t-2}$. Note that once $M1_{t-3}$ is larger than ${heaviside}(M0_{t-2},1)$ (Eq. \ref{equ:18}), this alpha becomes a formula without parameters. 
\small
\begin{align}
M:\quad \: S 1_{t} &= {std}(\min({heaviside}(M 0_{t-1},1),M 1_{t-2})) \label{equ:17} \\
P:\: M 1_{t-2} &= \max(M 0_{t-2},\min({heaviside}(M0_{t-2},1), M1_{t-3})) \label{equ:18} \\
U:\, M 1_{t-2} &= \max({transpose}(M0_{t-2}), \max(M0_{t-2}, \min \nonumber \\
&    ({heaviside}(M0_{t-2},1),M1_{t-3}))) \label{equ:19} 
% M 2_{t-2} &= {matrix\_uniform}(0.557673, 0.603385) \label{equ:20}
\end{align}
\normalsize
For $alpha\_AE\_B0\_4$ (Eq. \ref{equ:20} to Eq. \ref{equ:22}), the parameter $M 1_{t-2}$ is updated recursively with an expression of $M 0_{t-2}$ and an expression of $M 0_{t-4}$ (Eq. \ref{equ:21} and Eq. \ref{equ:22}). The prediction is based on the comparison between the inverse of ${close\_price}_{t-3}$ and the expression of $MV30_{t-4}$ (i.e., the moving average of the close prices over the last 30 days calculated at $t-4$), and the standard deviation of $M 1_{t-2}$ (Eq. \ref{equ:20}). Thus, this alpha makes trading decisions based on the recent close price ${close\_price}_{t-3}$, the trend of close prices $MV30_{t-4}$, and the volatility of the historically updated features $M 1_{t-2}$.
\small
\begin{align}
M:\quad \, S 1_{t} &= \tan(\tan(\min(1/{{close\_price}_{t-3}},\arcsin \label{equ:20} \\
& (\arcsin(MV30_{t-4})))))/std(M 1_{t-2}) \nonumber \\
U: M 1_{t-2} &= {matmul}(M2_{t-2}, M0_{t-2} M0_{t-2})  \label{equ:21} \\
M2_{t-2} &= {transpose}(\max({heaviside}({abs}( M2_{t-3})), \nonumber \\
& {broadcast}({vector\_uniform}(0.314561,-0.187581) \nonumber \\
& +{norm}(M0_{t-4},axis=0),axis=0))) \label{equ:22}
\end{align}
\normalsize
\begin{table}[t]
\captionof{table}{Ablation study of the parameter-updating function.}
\label{tab:Ablation Study Of Parameter-updating function}
\centering
\resizebox{\columnwidth}{!}{\begin{tabular}{ lrrr }  
\hline
\textbf{Alpha}  & \textbf{Sharpe ratio} & \textbf{IC} & \textbf{Correlation with the best alphas} \\
\hline
 alpha\_AE\_D\_0   & 21.323797&\textbf{0.067358}&NA\\
 alpha\_AE\_D\_0\_P   & \textbf{21.516798}&0.057707&NA\\
 \hline
 alpha\_AE\_R\_2  &\textbf{18.629571}&\textbf{0.066962}&-0.177144\\
 alpha\_AE\_R\_2\_P  &-0.344734&0.003149&-0.094286\\
 \hline
 alpha\_AE\_D\_3  &4.901069&\textbf{0.028437}&-0.202224\\
 alpha\_AE\_D\_3\_P   &\textbf{5.697408}&0.026347&-0.241651\\
 \hline
 alpha\_AE\_B0\_4 &\textbf{9.502871}&\textbf{0.032155}&0.137851\\
 alpha\_AE\_B0\_4\_P &-0.004294&-0.001908&-0.097541\\
\hline
\end{tabular}}
\end{table}
In Table \ref{tab:Ablation Study Of Parameter-updating function}, we perform the ablation study of the parameter-updating function, where each alpha without the parameter-updating function is denoted with an additional $P$.
This ablation study is crucial given the previous observation that the new alpha can be converted into a formulaic alpha.
We observe that all the parameter-updating functions are effective by increasing the fitness scores, i.e., ICs, in the inference stage. 
However, the Sharpe ratios do not change with the increasing ICs for $alpha\_AE\_D\_0\_P$ and $alpha\_AE\_D\_3\_P$.
This is expected because the Sharpe ratio of a portfolio depends on the top and bottom stock rankings while the IC measures rankings of all stocks. 

\begin{table}[t]
\tiny
\captionof{table}{Performance comparisons with the complex machine learning alphas.}
\label{tab:Comparison With Complex machine learning Models}
\centering
\resizebox{\columnwidth}{!}{\begin{tabular}{ lrrr }  
\hline
\textbf{Alpha}  & \textbf{Sharpe ratio} & \textbf{IC}\\
\hline
alpha\_AE\_D\_0   & \textbf{21.323797}& \textbf{0.067358}\\
alpha\_AE\_NN\_1   & 14.175835&0.065209\\
Rank\_LSTM   & 5.385036+/-1.608296& 0.027490+/-0.009336\\
RSR   & 5.647131+/-0.522782& 0.018623+/-0.000794\\
\hline
\end{tabular}}
\end{table}
\subsubsection{Comparisons With The Complex Machine Learning Alphas}
In Table \ref{tab:Comparison With Complex machine learning Models}, we compare the complex machine learning alphas with the generated alphas by AlphaEvolve. We observe that both the complex models fail to compete against $alpha\_AE\_D\_0$. 
The poor performance of RSR is due to the imposition of the relational domain knowledge to the data. 
The NASDAQ dataset cannot be best explained by the relational domain knowledge because the NASDAQ is a more noisy stock market compared to other stock markets (e.g., the NYSE)\footnote{https://finance.yahoo.com/news/nasdaq-vs-nyse-key-differences-200641822.html}: it incorporates many volatile stocks with less capitalization. 
Therefore, the noisy stock market affected by rapid-changing information cannot be modeled with the static relational knowledge. 
Besides, Rank\_LSTM and RSR are unstable with high standard deviations because their predictions are influenced by random effects (i.e., random seeds).
In contrast, the flexibility of the domain knowledge injection by AlphaEvolve leads to the best performance of $alpha\_AE\_D\_0$ without the domain knowledge, while $alpha\_AE\_NN\_1$ {generated} with the relational domain knowledge is weakly correlated with $alpha\_AE\_D\_0$ but has a {relatively} lower Sharpe ratio and IC.

\subsubsection{{Efficiency of The Pruning Technique}}
\label{Efficiency test for The Pruning Technique}
{In the ablation study of the pruning technique, we remove this technique from AlphaEvolve but use the prediction of an alpha as the fingerprint.
In Table \ref{tab:efficiency_test}, we denote each of the alphas without the technique with an additional letter $N$. The number of searched alphas is the sum of pruned alphas and evaluated alphas. We observe that, for the baseline methods, the numbers of searched alphas are significantly less than those of AlphaEvolve, leading to ineffective and inefficient alpha mining, which proves the effectiveness of our proposed technique.}
\begin{table}[t]
\captionof{table}{{Efficiency of the pruning technique.}}
\label{tab:efficiency_test}
\centering
\resizebox{\columnwidth}{!}{\begin{tabular}{ lrrrr }  
\hline
 \textbf{Alpha}  & \textbf{Sharpe ratio} & \textbf{IC} & \textbf{Correlation} & \textbf{Number of {searched} alphas} \\
\hline
 alpha\_AE\_D\_0   &   \textbf{21.323797}&\textbf{0.067358}&NA&309700\\
 alpha\_AE\_D\_0\_N  &8.898872&0.057817&NA&19500\\
 \hline
 alpha\_AE\_NN\_1   & \textbf{13.580572}&\textbf{0.056703}&-0.303845&1032700\\
 alpha\_AE\_NN\_1\_N  &5.148189&0.025506&-0.052586&5700\\  
 \hline
 alpha\_AE\_R\_2   &\textbf{18.629571}&\textbf{0.066962}&-0.177144&429800\\
 alpha\_AE\_R\_2\_N  &4.575985&0.032180&0.116183&13200\\

 \hline
 alpha\_AE\_D\_3   &\textbf{4.901069}&\textbf{0.028437}&-0.202224&910100\\
 alpha\_AE\_D\_3\_N  &4.604322&0.028945&-0.051935&37900\\ 

 \hline
 alpha\_AE\_B0\_4 &\textbf{9.502871}&\textbf{0.032155}&0.137851&220100\\
 alpha\_AE\_B0\_4\_N  &2.775825&0.027594&-0.137923&17300\\

\hline
\end{tabular}}
\end{table}
\section{Related Work}
\label{sec2}
Previous works focus on two {classes} of alphas: (1) machine learning alphas with vector and matrix operations; (2) formulaic alphas with scalar operations. For the first class, various models are proposed given more efficient deep learning systems to capture high-dimensional features \cite{10.1145/3003665.3003669}. \cite{10.1145/3318464.3389720} proposes a novel TITV architecture to model the time-invariant/variant features for the stock index prediction. \cite{10.5555/3172077.3172254} proposes a two-level attention mechanism to assign importance weights on time steps and stocks. \cite{zhang:SPP3097983.3098117} puts up a novel State Frequency Memory (SFM) that decomposes the hidden states of memory cells into multiple frequency components to model different trading activities. \cite{Ding:2015:DLE:2832415.2832572} uses a neural tensor network to extract event information from news data. \cite{DBLP:journals/corr/abs-1907-10046} encodes time-series images as candlestick (Box and Whisker) charts and models the image feature. \cite{Marko:book} creates a simple language of Japanese candlesticks {using} OHLC data. Domain knowledge injection has proven useful in designing alphas \cite{jiang2020applications, 10.1145/1376616.1376746}. 
\cite{feng2019temporal} injects relational knowledge through a novel graph network. {In general,} this {class} of alphas is designed by data scientists {and typically} too complex to {mine} into a weakly correlated set.

The second {class} of alphas is widely used in hedge funds. This {class} of alphas is designed by financial domain experts or mined by the genetic algorithm. \cite{RePEc:arx:papers:1601.00991} studies alphas invested in a hedge fund and their correlation standard. \cite{huatai} reports alphas mined by the genetic algorithm and their performance. \cite{huataiagain} further improves the genetic algorithm by using mutual information as the fitness score to mine nonlinear formulaic alphas. \cite{10.1145/191843.191925} proposes an indexing method to locate similar stock movement patterns. \cite{Lu1998StockMP} proposes N-dimensional inter-transaction rules to predict stock movements. To our best knowledge, our work is the first to mine alphas {based on} AutoML and first to {generate} the new {class} of alphas {combining the strengths of the existing classes}.

\section{Conclusions}
\label{sec5}

In this paper, we {introduce a new class of alphas and then} propose a novel alpha mining framework AlphaEvolve based on AutoML. AlphaEvolve {generates the new class of} alphas which are different from previous {classes} of formulas and machine learning models. This {class} has the advantages of simplicity and generalization ability similar to formulaic alphas, and the ability to be trained by data similar to machine learning alphas. These advantages {result in} better performances in {generating} weakly correlated high returns. {Consequently, AlphaEvolve provides investors with an automatic solution for low-risk investments with high returns.}

\section{Acknowledgement}
This research is funded by Singapore Ministry of Education Academic Research Fund Tier 3 with the award number MOE2017-T3-1-007. Meihui Zhang's research is supported by National Natural Science Foundation of China (62050099).

\bigskip

\bibliographystyle{ACM-Reference-Format}
\bibliography{reference}

\end{document}